\def\year{2016}\relax
\documentclass[letterpaper]{article}
\usepackage{aaai16}
\usepackage{natbib}
\usepackage{mathptmx}
\usepackage{times}
\usepackage{helvet}
\usepackage{courier}
\usepackage{algorithm}
\usepackage[noend]{algorithmic}
\usepackage{graphicx}
\usepackage{xcolor}
\usepackage{float}
\usepackage{amsmath}
\usepackage{amssymb}
\usepackage{amsthm}
\usepackage{subfig}
\usepackage{booktabs}

\newtheorem{exampl}{Example}
\newenvironment{example}{\begin{exampl}\em}{\end{exampl}}
\newtheorem{defn}{Definition}
\newenvironment{definition}{\begin{defn}\em}{\end{defn}}

\newcommand{\sigmoid}{\sigma}

\begin{document}

\title{A Learning Algorithm for Relational Logistic Regression: Preliminary Results\thanks{In IJCAI-16 Statistical Relational AI Workshop.}}
\author{
Bahare Fatemi, Seyed Mehran Kazemi \and David Poole\\
The University of British Columbia\\
Vancouver, BC, V6T 1Z4\\
\{bfatemi, smkazemi, poole\}@cs.ubc.ca
}
\maketitle
\begin{abstract}
Relational logistic regression (RLR) is a representation of conditional probability in terms of weighted formulae for modelling multi-relational data. In this paper, we develop a learning algorithm for RLR models. Learning an RLR model from data consists of two steps: 1- learning the set of formulae to be used in the model (a.k.a. structure learning) and learning the weight of each formula (a.k.a. parameter learning). For structure learning, we deploy Schmidt and Murphy's hierarchical assumption: first we learn a model with simple formulae, then more complex formulae are added iteratively only if all their sub-formulae have proven effective in previous learned models. For parameter learning, we convert the problem into a non-relational learning problem and use an off-the-shelf logistic regression learning algorithm from Weka, an open-source machine learning tool, to learn the weights. We also indicate how hidden features about the individuals can be incorporated into RLR to boost the learning performance. We compare our learning algorithm to other structure and parameter learning algorithms in the literature, and compare the performance of RLR models to standard logistic regression and RDN-Boost on a modified version of the MovieLens data-set.

\end{abstract}

Statistical relational learning (SRL) \citep{StarAI-Book} aims at unifying logic and probability to provide models that can learn from complex multi-relational data. Relational probability models (RPMs) \citep{Getoor:2007} (also called template-based models \citep{Koller:2009}) are the core of the SRL systems. They extend Bayesian networks and Markov networks \citep{Pearl:1988} by adding the concepts of objects, properties and relations, and by allowing for probabilistic dependencies among relations of individuals. 

Unlike Bayesian networks, in RPMs a random variable may depend on an unbounded number of parents in the grounding. In these cases, the conditional probability cannot be represented as a table. To address this issue, many of the existing relational models (e.g., \citep{DeRaedt:2007,natarajan2012gradient}) use simple aggregation models such as existential quantifiers, or noisy-or models. These models are much more compact than tabular representations. There also exists other aggregators with different properties (e.g., see \citep{reason:HorPoo90a,Friedman:1999,neville05,Perlich:2006,Kisynski:2009,Natarajan:2010aa}).

Relational logistic regression (RLR) \citep{Kazemi:2014} has been recently proposed as an aggregation model which can represent much more complex functions than the previous aggregators. RLR uses weighted formulae to define a conditional probability. Learning an RLR model from data consists of a structure learning and a parameter learning phase. The former corresponds to learning the features (weighted formulae) to be included in the model, and the latter corresponds to learning the weight of each feature. When all of the parents are observed (e.g., for classification), \citet{Poole:2014} observed that an RLR model has similar semantics as a Markov logic network (MLN) \citep{Richardson:2006aa}. Therefore, one can use a discriminative learning algorithm for MLNs to learn an RLR model. 

\citet{Huynh:2008} proposed a bottom-up algorithm for discriminative learning of MLNs. They use a logic program learner (ALEPH \citep{srinivasan2001aleph}) to learn the structure, and then use L1-regularized logistic regression to learn the weights and enable automatic feature selection. The problem with this approach is that the ALEPH (or any other logic program learner) and the MLN have different semantics: features marked is useful by ALEPH are not necessarily useful features for an MLN, and features marked as useless by ALEPH are not necessarily useless features for MLN. The former happens because ALEPH generates features with logical accuracy which is only a rough estimate of their value in a probabilistic model. The latter happens because the representational power of an MLN is more than that of ALEPH: MLNs can leverage counts but ALEPH is based on the existential quantifier. While the former issue may be resolved using L1-regularization, it is not straight-forward to resolve the latter issue.

In this paper, we develop and test an algorithm for learning RLR models from relational data which addresses the aforementioned issues with current learning algorithms. 
Our learning algorithm follows the hierarchical assumption of \citep{Schmidt:2010}: a formula $a \wedge b$ may be a useful feature only if $a$ and $b$ have each proven useful. Our learning algorithm is, in spirit, similar to the refinement graphs of \citet{popescul2004cluster}. In our algorithm, however, feature generation is done using hierarchical assumption instead of query refinement, and feature selection is done for each level of hierarchy using hierarchical (or L1) regularization instead of sequential feature selection, thus reducing the number of times a model is learned from data and enabling previously selected features to be removed as the search proceeds.

We also incorporate hidden features about each individual in our RLR model and observe how they affect the performance. We test our learning algorithm on a modified version of the MovieLens data-set taken from \citet{schulte2012learning} for predicting users' gender and age. We compare our model with standard logistic regression models that do not use relational features, as well as the RDN-Boost \citep{natarajan2012gradient} which is one of the state-of-the-art relational learning algorithms. The obtained results show that RLR can learn more accurate models compared to the standard logistic regression and RDN-Boost. The results also show that adding hidden features to RLR may increase the accuracy of the predictions, but may make the model over-confident about its predictions. Regularizing the RLR predictions towards the mean of the data helps avoid over-confidency.

\section{Background and Notations}
In this section, we introduce the notation used throughout the paper, and provide necessary background information for readers to follow the rest of the paper.

\subsection{Logistic Regression}
Logistic regression (LR) \citep{Allison:1999} is a popular classification method within machine learning community. We describe how it can be used for classification following \citet{Cessie:1992} and \citet{Mitchell:1997}. 

Suppose we have a set of labeled examples $\{(x_1, y_1), (x_2, y_2), \dots, (x_m, y_m)\}$, where each $x_i$ is composed of $n$ features $x_{i1}, x_{i2}, \dots, x_{in}$ and each $y_i$ is a binary variable whose value is to predicted. The $x_{ij}$s may be binary, multi-valued or continuous. Throughout the paper, we assume binary variables take their values from $\{0, 1\}$. Logistic regression learns a set $w=\{w_0, w_1, \dots, w_n\}$ of weights, where $w_{0}$ is the intercept and $w_j$ is the weight of the feature $x_{ij}$. For simplicity, we assume a new dimension $x_{i0} = 1$ has been added to the data to avoid treating $w_0$ differently than other $w_j$s. LR defines the probability of $y_i$ being $1$ given $x_i$ as follows:
\begin{equation}
\label{lr}
P(y_i = 1 \mid x_i, w) = \sigmoid(\sum_{j=0}^{n} x_{ij}w_j) \\
\end{equation}
where $\sigmoid(x)=\frac{1}{1+exp(-x)}$ is the Sigmoid function.

Logistic regression learns the weights by maximizing the log-likelihood of the data (or equivalently, minimizing the logistic loss function) as follows: 
\begin{equation}
w_{LR} = argmax_w \sum_{i=0}^{m} log(P(y_i = 1 \mid x_i))
\end{equation}
An L1-regularization can be added to the loss function to encourage sparsity and do an automatic feature selection.

\subsection{Conjoined Features and the Hierarchical Assumption}
Given $n$ input random variables, logistic regression considers $n+1$ features: one bias (intercept) and one feature for each random variable. One can generate more features by conjoining the input random variables. For instance, if $a$ and $b$ are two continuous random variables, one can generate a new feature $a * b$ (which is $a \wedge b$ for Boolean variables). Given $n$ random variables, conjoining (or multiplying) variables allows for generating $2^n$ features. These $2^n$ weights can represent arbitrary conditional probabilities\footnote{Note that there are $2^n$ degrees of freedom and any representation that can represent arbitrary conditional probabilities may require $2^n$ parameters. The challenge is to find a representation that can often use fewer.}, in what is known as the canonical representation - refer to \citet{Buchman:2012} or \citet{Koller:2009}. For a large $n$, however, generating $2^n$ features may not be practically possible, and it also makes the model overfit to the training data.  

In order to avoid generating all $2^n$ features, \citet{Schmidt:2010} make a hierarchical assumption: if either $a$ or $b$ are not useful features, neither is $a \wedge b$. Having this assumption, \citet{Schmidt:2010} first learn an LR model considering only features with no conjunctions. They regularize 
their loss function with a hierarchical regularization function, so that the weights of the features not contributing to the prediction go to zero. Once the learning stops, they keep the features having non-zero weights, and add all conjoined features whose all subsets have non-zero weights in the previous step. Then they run their learning again. They continue this process until no more features can be added.

\subsection{Relational Logistic Regression}
Relational logistic regression (RLR) \citep{Kazemi:2014} is the analogue of LR for relational models. RLR can be also considered as the directed analogue of Markov logic networks \citep{Richardson:2006aa}. In order to describe RLR, first we need to introduce some definitions and terminologies used in relational domains.

A {\bf population} refers to a set of {\bf individuals} and corresponds to a domain in logic. {\bf Population size} of a population is a non-negative number indicating its cardinality. For example, a population can be the set of planets in the solar system, where {\it Mars} is an individual and the population size is {\it 8}.

\textbf{Logical variables} start with lower-case letters, and \textbf{constants} start with upper-case letters.  Associated with a logical variable $x$ is a population $pop(x)$ where $|x| = |pop(x)|$ is the size of the population. A lower-case and an upper-case letter written in bold refer to a set of logical variables and a set of individuals respectively.

A {\bf parametrized random variable (PRV)} \citep{Poole:2003} is of the form $F(t_1, ..., t_k)$  where $F$ is a k-ary (continuous or categorical) function symbol and each $t_i$ is a logical variable or a constant. If all $t_i$s are constants, the PRV is a random variable. If k = 0, we can omit the parentheses. If $F$ is a predicate symbol, $F$ has range \{True, False\}, otherwise the range of $F$ is the range of the function. For example, $LifeExistsOn(planet)$ can be a PRV with predicate function $LifeExistsOn$ and logical variable $planet$, which is true if life exists on the given $planet$.

A {\bf literal} is an assignment of a value to a PRV. We represent $F(.)=true$ as $f(.)$ and $F(.)=false$ as $\neg f(.)$. A \textbf{formula} is made up of literals connected with conjunction or disjunction.

\begin{figure}
\begin{center}
\includegraphics[width=0.5\columnwidth]{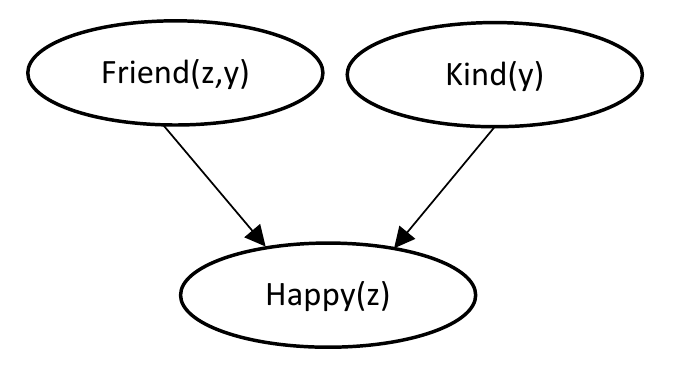}
\end{center}
\caption{A relational model taken from \citep{Kazemi:2014}.}
\label{KindFriend}
\end{figure}


A {\bf weighted formula (WF)} for a PRV $Q(\mathbf{z})$, where $\mathbf{z}$ is a set of logical variables, is a tuple $\left\langle F,w \right\rangle$ where $F$ is a Boolean formula of the parents of $Q$ and $w$ is a weight. 

{\bf Relational logistic regression (RLR)} defines a conditional probability distribution for a Boolean PRV $Q(\mathbf{z})$ using a set of WFs $\psi$ as follows:
\begin{equation}
P(q(\mathbf{Z}) \mid \Pi) = \sigmoid\big(\sum_{\left\langle F,w \right\rangle \in \psi}{w*F_{\Pi,\mathbf{z}\rightarrow \mathbf{Z}}}\big)
\end{equation}

where $\sigmoid(x)=\frac{1}{1+exp(-x)}$ is the Sigmoid function, $\Pi$ represents the assigned values to parents of $Q$, $\mathbf{Z}$ represents an assignment of individuals to the logical variables in $\mathbf{z}$, and $F_{\Pi,\mathbf{z}\rightarrow \mathbf{Z}}$ is formula $F$ with each logical variable $\mathbf{z}$ in it being replaced according to $\mathbf{Z}$, and evaluated in $\Pi$.


\begin{example}
Consider the relational model in Fig. \ref{KindFriend} taken from \citep{Kazemi:2014} and suppose we want to model \texttt{"}someone is happy if they have at least 5 friends that are kind\texttt{"}. The following WFs can be used to represent this model:
\[\begin{array}{c}
\mbox{$\left\langle True,-4.5 \right\rangle$}\\
\mbox{$\left\langle friend(z,y) \wedge kind(y),1 \right\rangle$} \\
\end{array}\] 
RLR sums over the above WFs resulting in:
\begin{center}
$\forall Z \in z: P(Happy(Z)=True | Friend(Z,y),Kind(y)) = Sigmoid(-4.5 + 1*n_T)$ 
\end{center}
where $n_T$ represents the number of individuals in $y$ for which $Friend(Z,y) \wedge Kind(y)$ is true. When $n_T \geq 5$, the probability is closer to one than zero and when $n_T < 5$, the probability is closer to zero than one.
\end{example}

\subsection{Handling Continuous Variables}
RLR was initially designed for Boolean or multi-valued parents. If $True$ is associated with $1$ and $False$ is associated with $0$, we can substitute $\wedge$ in our WFs with $*$. Then we can allow continuous PRVs in WFs. For example if for some $X \in x$ we have $R(X)=True$, $S(X)=False$ and $T(X)=0.2$, then $\left\langle r(X) * s(X),w \right\rangle $ evaluates to $0$, $\left\langle r(X) * \neg s(X),w \right\rangle$ evaluates to $w$, and $\left\langle r(X) * s(X) * t(X),w \right\rangle $ evaluates to $0.2*w$.



\section{Learning Relational Logistic Regression}
The aforementioned learning algorithm for LR is not directly applicable to RLR because the number of potential weights in RLR is unbounded. We develop a learning algorithm for RLR which can handle the unbounded number of potential weights.

Learning RLR from data consists of two parts: structure learning (learning the set of WFs) and parameter learning (learning the weight of each WF).
\subsection{Parameter Learning}
Suppose we are given a set $\psi$ of WFs defining the conditional probability of a PRV $Q(\mathbf{z})$, and we want to learn the weight of each WF. We can convert this learning problem into a LR learning problem by generating a flat data-set in single-matrix form. To do so, for each assignment $\mathbf{Z}$ of individuals to the logical variables in $\mathbf{z}$, we generate one data row $\left\langle x_{i1}, x_{i2}, \dots, x_{i|\psi|}, y_i \right\rangle$ in which $y_i = Q(\mathbf{Z})$ and $x_{ij}$ is the number of times the formula of the $j$-th WF in $\psi$ is true when the logical variables in $\mathbf{z}$ are replaced with the individuals in $\mathbf{Z}$. Once we do this conversion, we have a single-matrix data for which we can learn a LR model. The weight learned for the $j$-th input gives the weight for our $j$-th WF. Note that the conversion is complete and specifies the values of all relations.

\begin{example}
Consider the relational model in Fig.~\ref{KindFriend} and suppose we are given the following WFs:
\begin{center}
\mbox{$\left\langle True,w_1 \right\rangle$}\\
\mbox{$\left\langle friend(z,y) ,w_2 \right\rangle$} \\
\mbox{$\left\langle friend(z,y) * kind(y),w_3 \right\rangle$} \\
\end{center}
In this case, we generate a matrix data having a row for each individual $Z \in z$ where the row consists of four values: the first number is $1$ serving as the intercept, the second one is the number of people that are friends with $Z$, the third is the number of kind people that are friends with $Z$, and the fourth one represents whether $Z$ is happy or not. These are sufficient statistics for learning the weights. An example of the generated matrices is as follows:
\begin{table}[h]
  \label{results}
  \centering
  \begin{tabular}{ccc|c}
    Bias & \#Friends & \#Kind Friends & Happy? \\
    \midrule
    1 & 5 & 3 & Yes \\
    1 & 18 & 2 & No \\
    1 & 1  & 1 & Yes \\
    1 & 12 & 10 & Yes \\
    \dots & \dots & \dots & \dots
  \end{tabular}
\end{table}

In the above matrix, the first four people have $5$, $18$, $1$ and $12$ friends respectively, out of which $3$, $2$, $1$, and $10$ are kind. 
\end{example}

\subsection{Structure Learning}
Learning the structure of an RLR model refers to selecting a set of WFs that should be used. By conjoining different relations and adding attributes, one can generate an infinite number of WFs. As an example, suppose we want to predict the gender of users ($G(u)$) in a movie rating system, where we are given the occupation ($O(u)$), age ($A(u)$) and the movies that these users have rated ($Rated(u,m)$), as well as the set of (possibly more than one) genres that each movie belongs to. Our WFs can have formulae such as:
\begin{center}
$a(u)=young$\\ 
$rated(u, m) * drama(m)$ \\
$rated(u, m) * rated(u', m) * g(u')=male$ 
\end{center}
and many other WFs with much more conjoined relations. Not all of these WFs may be useful though. As an example, a WF whose formula is $action(m)$ is not a useful feature in predicting $G(u)$ as it evaluates to a constant number for all users. We avoid generating such features in an RLR learning model. To do this in a systematic way, we need a few definitions:

\begin{definition}
Let $\psi$ denote the set of WFs defining the conditional probability of a PRV $Q(\mathbf{z})$. A logical variable $v$ in a formula $f$ of a WF $\in \psi$ is a:
\begin{itemize}
\item \textbf{target} if $v \in \mathbf{z}$
\item \textbf{connector} for $v_1$ and $v_2$ if there are at least two relations in $f$ one having $v$ and $v_1$ and the other having $v$ and $v_2$
\item \textbf{attributed} if there exists at least one PRV in $f$ having only $v$ (e.g., $g(v)$)
\item \textbf{hanging} if it fits in none of the above definitions
\end{itemize}
\end{definition}
\begin{definition}
A WF is a \emph{chain} \citep{schulte2012learning} if its literals can be ordered as a list $[r_1(\mathbf{x_1}), \dots, r_k(\mathbf{x_k})]$ such that each literal $r_{i+1}(\mathbf{x_{i+1}})$ shares at least one logical variable with the preceding literals $r_1(\mathbf{x_1}), \dots, r_i(\mathbf{x_i})$. A chain is \emph{targeted} if it has at least one target logical variable. A WF is \emph{k-BL} if it contains no more than $k$ binary literals, and is \emph{r-UL} if it contains no more than $r$ unary literals.
\end{definition}

\begin{example}
Suppose \\
$\left\langle rated(u,m) * rated(u',m) * rated(u',m') * comedy(m),w \right\rangle$ \\
is a WF belonging to the set of WFs defining the conditional probability of $G(u)$. Then $u$ is a target, $m$ and $u'$ are connectors, $m$ is also attributed, and $m'$ is a hanging logical variable. This WF is a chain because the second literal shares a $m$ with the first literal, the third one shares a $u'$ with the second one, and the fourth one shares a $m$ with the first and second literal. This chain is targeted because it contains $u$, which is the only target logical variable. The WF is $3$-$BL$, $1$-$UL$ as it contains no more than $3$ binary and no more than $1$ unary literals.
$\left\langle age(u)=young * comedy(m), w \right\rangle$ is a non-chain WF, and $\left\langle drama(m) * comedy(m), w \right\rangle$ is a chain which is not targeted. 
\end{example}

We avoid generating two types of WFs: 1- WFs that are not targeted chains, 2- WFs that contain hanging logical variables. We do this because non-targeted chains (e.g., $\left\langle drama(m) * comedy(m), w \right\rangle$ for predicting $G(u)$) always evaluates to a constant number, and WFs with hanging logical variables (e.g., $\left\langle rated(u, m) * acted(a, m), w \right\rangle$ for predicting $G(u)$) can be replaced with more informative WFs. 
In the rest of the paper, we only consider the WFs that are targeted chains and have no hanging logical variables.

Having these definitions, we state the hierarchical assumption as follows: \\
\textbf{Hierarchical Assumption:} Let $f$ be a $k$-$BL$, $r$-$UL$ WF. Let $\psi_f$ be the set of all $k$-$BL$, $j$-$UL$ ($j < r$) WFs having the same $k$ binary literals and a strict subset of the unary literals as $f$. $f$ is \emph{useless} if L1-regularized logistic regression assigns a zero weight to it, or there exists a useless WF in $\psi_f$.
\begin{example}
A WF $\left\langle rated(u, m) * drama(m) * comedy(m), w \right\rangle$ is useless if either $\left\langle rated(u, m) * drama(m), w_1 \right\rangle$ or $\left\langle rated(u, m) * comedy(m), w_2\right\rangle$ is useless, or L1-regularization sets $w$ to 0.
\end{example}

In order to learn the structure of an RLR model, we select a value $k$ and generate all allowed \emph{k-BL,1-UL} WFs. We find the best value of $k$ by cross-validation. Then we add WFs with more unary literals by making the hierarchical assumption and using a similar search strategy as in \citep{Schmidt:2010} by following the algorithm below: \\
$curWFs \leftarrow $ $set$ $of$ $k$-$BL$, $1$-$UL$ $WFs$\\
$removedWFs \leftarrow \emptyset$ \\
$r \leftarrow 1$ \\
$while(stoppingCriteriaMet())\{$ \\
\indent $\mathbb{M}=L1$-$Regularized$-$Logistic$-$Regression(curWFs)$ \\
\indent $removedWFs$ += $curWFs$ $having$ $w=0$ $in$ $\mathbb{M}$\\
\indent $r \leftarrow r + 1$ \\
\indent $curWFs$ = $HA(removedWFs,r)$ \\
$\}$

The algorithm starts with $k$-$BL$, $1$-$UL$ WFs. Initially, no WF is labeled as removed. $r$ is initially set to $1$ to indicate the current maximum number of unary literals. Then until the stopping criteria is met, the weights of the WFs in $curWFs$ are learned using an L1-regularized logistic regression. If the weight of a WF is set to zero, we add it to the $removedWFs$. Then we increment $r$ and update the $curWFs$ to the $k$-$BL$, $r$-$UL$ WFs obeying the hierarchical assumption with respect to the $removedWFs$ (we assume $HA(removedWFs,r)$ is a function which returns such WFs). The stopping criteria is met when no more WFs can be generated as $r$ increases.

\subsection{Adding Hidden Features}
While we exploit the observed features of the objects in making predictions, each object may contain really useful information that has not been observed. As an example, in predicting the gender of the users given the movies they liked, some movies may only be appealing to males and some only to females. Or there might be features in movies that we do not know about, but they contribute to predicting the gender of users.

In order to incorporate hidden features in our RLR model, we add continuous unary PRVs such as $H(m)$ with (initially) random values to our dataset. Then we generate all $k$-$BL$, $1$-$UL$ WFs and learn the weights as well as the values of the hidden features using stochastic gradient descent with L1-regularization. Once we learn the values of the hidden features, we treat them as normal features and use our aforementioned structure and parameter learning algorithms to learn an RLR model.

\begin{table*}[t]
  \caption{ACLL and accuracy on predicting the gender and age of the users in the MovieLens dataset.}
  \label{results}
  \centering
  \begin{tabular}{ccccccc}
    \toprule
    & & \multicolumn{5}{c}{Learning Algorithm}                   \\
    \cmidrule{3-7}
                   &                                                         & Baseline        &  LR               & RDN-Boost    & RLR-Base       & RLR-H                 \\
    \midrule
	Gender	  & \multicolumn{1}{c|}{ACLL}            &  -0.6020        &   -0.5694       &    -0.5947        &   -0.5368        &  -0.5046\\
		        & \multicolumn{1}{c|}{Accuracy}       &  71.0638\%   &   71.383\%    &    70.6383\%   &   73.8298\%   &  77.3404\% \\
		    
       \midrule 		    
    
          Age      & \multicolumn{1}{c|}{ACLL}             &  -0.6733        &   -0.5242       &   -0.5299        &  -0.5166        & -0.5090\\
                      & \multicolumn{1}{c|}{Accuracy}       &  60.1064\%   &   76.0638\%   &   76.4893\%   &  77.1277\%   & 77.0212\%\\
              
    \bottomrule
  \end{tabular}
\end{table*}

\section{Experiments and Results}
We test our learning algorithm on the 0.1M Movielens data-set \citep{harper2015movielens} with the modifications made by \citet{schulte2012learning}. This data-set contains information about $940$ users (nominal variables for age, occupation, and gender), $1682$ movies (binary variables for action, horror, and drama), the movies rated by each user containing $79,778$ user-movie pairs, and the actual rating the user has given to a movie. In our experiments, we ignored the actual ratings and only considered if a movie has been rated by a user or not. 

We learned RLR models for predicting the age and gender of users once with no hidden features (RLR-Base), and once with one hidden feature (RLR-H) for the movies. We regularize the predictions of both RLR-Base and RLR-H towards the mean as:\\
\begin{equation}
Probability = \lambda * mean + (1 - \lambda) * (RLR signal)
\end{equation}
When predicting the age of the users, we only considered two instead of three age classes (we merged the age\_1 and age\_2 classes). 
For learning Logistic regression models with L1-regularization, we used the open source codes of \citep{schmidt2007fast} and we learned the final logistic regression model with Weka software \citep{hall2009weka}. 
We compared the proposed method with a baseline model always predicting the mean, standard logistic regression (LR) not using the relational information, and the RDN-Boost. 

The performance of all learning algorithms were obtained by 5-folds cross-validation. In each fold, we divided the users in the Movielens data-set randomly into 80\% training set and 20\% test set. We learned the model on the train set and measured the accuracy (the percentage of correctly classified instances) and the average conditional log-likelihood (ACLL) on the test set, and averaged them over the 5 folds. ACLL is computed as follows:
\begin{equation}
ACLL = \frac{1}{m} \sum_{i=1}^m ln(P(G(U_i) \mid data, model))
\end{equation}

Obtained results are represented in Table~\ref{results}. They show that RLR utilizes the relational features to improve the predictions compared to the logistic regression model that does not use the relational information, and the RDB-Boost. 
Obtained results also represent that adding hidden features to the RLR models may increase the accuracy and reduce the MAE. However, we observed that adding hidden features makes the model over-confident by pushing the prediction probabilities towards zero and one, thus requiring more regularization towards the mean. 

\section{Discussion}
In our first experiment on predicting the gender of the users, we found that on average men have rated more action movies than women. This means for predicting the gender of the users, the feature $male(u) \Leftarrow rated(u, m) \wedge action(m)$ is a useful feature for both RLR and MLNs as it counts the number of action movies. Many of the current relational learning algorithms/models, however, rely mostly on the existential quantifier as their aggregator (e.g., \citep{reason:HorPoo90a,DeRaedt:2007,Huynh:2008,natarajan2012gradient}). By relying on the existential quantifier, these models either have to use many complex rules to imitate the effect of such rules, or lose great amounts of relational information available in terms of counts.

As a particular example, consider the discriminative structure learning of \citet{Huynh:2008} for MLNs. First they learn a large set of features using ALEPH, then learn the weights of the features with L1-regularization to enable automatic feature selection. In cases where everyone has rated an action movie, $male(u)$ :- $rated(u, m) \wedge action(m)$ is not a useful feature for ALEPH (and so it will not find it) because it does not distinguish males from females. Therefore, this rule will not be included in the final MLN. 

Relational learners based on existential quantifier can potentially imitate the effect of counts by using many complex rules. As an example, $male(u)$ :- $rated(u, m_1) \wedge action(m_1) \wedge rated(u, m_2) \wedge action(m_2) \wedge m_1 \neq m_2$ may be used to assign a maleness probability to people rating two action movies. But this approach requires a different rule for each count, and the rules become more and more complex as the count grows because they require pairwise inequalities. To see this in practice, we ran experiments with ALEPH on synthesized data and observed that, even though it could learn such rules to enhance its predictions, it failed at finding them.

Based on the above observations, we argue that relational learning algorithms/models need to allow for a richer set of predefined aggregators, or enable non-predefined aggregators to be learned from data similar to what RLR does. We also argue that our structure learning algorithm has the potential to explore more features, and may also be a good candidate for discriminative structure learning of MLNs.

\section{Conclusion}
Relational logistic regression (RLR) can learn complex models for multi-relational data-sets. In this paper, we developed and tested a structure and parameter learning for these models based on the hierarchical assumption. We compared our model with the standard logistic regression model and the RDN-Boost, and represented that, on the MovieLens data-set, RLR achieves higher accuracies. We also represented how hidden features can boost the performance of RLR models. The results presented in this work are only preliminary results.
Future direction includes testing our learning algorithm on more complex data-sets having much more relational information, comparing our model with other relational learning and aggregation models in the literature, making the learning algorithm extrapolate properly for the un-seen population sizes \citep{Poole:2014}, and testing the performance of our structure learning algorithm for discriminative learning of Markov logic networks.

\bibliography{MyBib}

\begin{thebibliography}{}

\bibitem[\protect\citeauthoryear{Allison}{1999}]{Allison:1999}
Allison, P.
\newblock 1999.
\newblock {\em Logistic regression using SAS®: theory and application}.
\newblock SAS Publishing.

\bibitem[\protect\citeauthoryear{Buchman \bgroup et al\mbox.\egroup
  }{2012}]{Buchman:2012}
Buchman, D.; Schmidt, M.; Mohamed, S.; Poole, D.; and {De Freitas}, N.
\newblock 2012.
\newblock On sparse, spectral and other parameterizations of binary
  probabilistic models.
\newblock In {\em AISTATS}.

\bibitem[\protect\citeauthoryear{Cessie and van
  Houwelingen}{1992}]{Cessie:1992}
Cessie, S.~L., and van Houwelingen, J.
\newblock 1992.
\newblock Ridge estimators in logistic regression.
\newblock {\em Applied Statistics} 41(1):191--201.

\bibitem[\protect\citeauthoryear{{De Raedt} \bgroup et al\mbox.\egroup
  }{2016}]{StarAI-Book}
{De Raedt}, L.; Kersting, K.; Natarajan, S.; and Poole, D.
\newblock 2016.
\newblock Statistical relational artificial intelligence: Logic, probability,
  and computation.
\newblock {\em Synthesis Lectures on Artificial Intelligence and Machine
  Learning} 10(2):1--189.

\bibitem[\protect\citeauthoryear{{De Raedt}, Kimmig, and
  Toivonen}{2007}]{DeRaedt:2007}
{De Raedt}, L.; Kimmig, A.; and Toivonen, H.
\newblock 2007.
\newblock Problog: A probabilistic prolog and its application in link
  discovery.
\newblock In {\em IJCAI}, volume~7.

\bibitem[\protect\citeauthoryear{Friedman \bgroup et al\mbox.\egroup
  }{1999}]{Friedman:1999}
Friedman, N.; Getoor, L.; Koller, D.; and Pfeffer, A.
\newblock 1999.
\newblock Learning probabilistic relational models.
\newblock In {\em Proc. of the Sixteenth International Joint Conference on
  Artificial Intelligence},  1300--1307.
\newblock Sweden: Morgan Kaufmann.

\bibitem[\protect\citeauthoryear{Getoor and Taskar}{2007}]{Getoor:2007}
Getoor, L., and Taskar, B.
\newblock 2007.
\newblock {\em Introduction to Statistical Relational Learning}.
\newblock MIT Press, Cambridge, MA.

\bibitem[\protect\citeauthoryear{Hall \bgroup et al\mbox.\egroup
  }{2009}]{hall2009weka}
Hall, M.; Frank, E.; Holmes, G.; Pfahringer, B.; Reutemann, P.; and Witten,
  I.~H.
\newblock 2009.
\newblock The weka data mining software: an update.
\newblock {\em ACM SIGKDD explorations newsletter} 11(1):10--18.

\bibitem[\protect\citeauthoryear{Harper and
  Konstan}{2015}]{harper2015movielens}
Harper, M., and Konstan, J.
\newblock 2015.
\newblock The movielens datasets: History and context.
\newblock {\em ACM Transactions on Interactive Intelligent Systems (TiiS)}
  5(4):19.

\bibitem[\protect\citeauthoryear{Horsch and Poole}{1990}]{reason:HorPoo90a}
Horsch, M., and Poole, D.
\newblock 1990.
\newblock A dynamic approach to probability inference using {B}ayesian
  networks.
\newblock In {\em Proc. sixth Conference on Uncertainty in AI},  155--161.

\bibitem[\protect\citeauthoryear{Huynh and Mooney}{2008}]{Huynh:2008}
Huynh, T.~N., and Mooney, R.~J.
\newblock 2008.
\newblock Discriminative structure and parameter learning for markov logic
  networks.
\newblock In {\em Proc. of the international conference on machine learning}.

\bibitem[\protect\citeauthoryear{Kazemi \bgroup et al\mbox.\egroup
  }{2014}]{Kazemi:2014}
Kazemi, S.~M.; Buchman, D.; Kersting, K.; Natarajan, S.; and Poole, D.
\newblock 2014.
\newblock Relational logistic regression.
\newblock In {\em Proc. 14th International Conference on Principles of
  Knowledge Representation and Reasoning (KR)}.

\bibitem[\protect\citeauthoryear{Kisynski and Poole}{2009}]{Kisynski:2009}
Kisynski, J., and Poole, D.
\newblock 2009.
\newblock Lifted aggregation in directed first-order probabilistic models.
\newblock In {\em Twenty-first International Joint Conference on Artificial
  Intelligence},  1922--1929.

\bibitem[\protect\citeauthoryear{Koller and Friedman}{2009}]{Koller:2009}
Koller, D., and Friedman, N.
\newblock 2009.
\newblock {\em Probabilistic Graphical Models: Principles and Techniques}.
\newblock MIT Press, Cambridge, MA.

\bibitem[\protect\citeauthoryear{Mitchell}{1997}]{Mitchell:1997}
Mitchell, T.
\newblock 1997.
\newblock {\em Machine Learning}.
\newblock McGraw Hill.

\bibitem[\protect\citeauthoryear{Natarajan \bgroup et al\mbox.\egroup
  }{2010}]{Natarajan:2010aa}
Natarajan, S.; Khot, T.; Lowd, D.; Tadepalli, P.; and Kersting, K.
\newblock 2010.
\newblock Exploiting causal independence in {M}arkov logic networks: Combining
  undirected and directed models.
\newblock In {\em European Conference on Machine Learning (ECML)}.

\bibitem[\protect\citeauthoryear{Natarajan \bgroup et al\mbox.\egroup
  }{2012}]{natarajan2012gradient}
Natarajan, S.; Khot, T.; Kersting, K.; Gutmann, B.; and Shavlik, J.
\newblock 2012.
\newblock Gradient-based boosting for statistical relational learning: The
  relational dependency network case.
\newblock {\em Machine Learning} 86(1):25--56.

\bibitem[\protect\citeauthoryear{Neville \bgroup et al\mbox.\egroup
  }{2005}]{neville05}
Neville, J.; Simsek, O.; Jensen, D.; Komoroske, J.; Palmer, K.; and Goldberg,
  H.
\newblock 2005.
\newblock Using relational knowledge discovery to prevent securities fraud.
\newblock In {\em Proceedings of the 11th ACM SIGKDD International Conference
  on Knowledge Discovery and Data Mining}.
\newblock MIT Press.

\bibitem[\protect\citeauthoryear{Pearl}{1988}]{Pearl:1988}
Pearl, J.
\newblock 1988.
\newblock {\em Probabilistic Reasoning in Intelligent Systems: Networks of
  Plausible Inference}.
\newblock San Mateo, CA: Morgan Kaumann.

\bibitem[\protect\citeauthoryear{Perlish and Provost}{2006}]{Perlich:2006}
Perlish, C., and Provost, F.
\newblock 2006.
\newblock Distribution-based aggregation for relational learning with
  identifier attributes.
\newblock {\em Machine Learning} 62:65--105.

\bibitem[\protect\citeauthoryear{Poole \bgroup et al\mbox.\egroup
  }{2014}]{Poole:2014}
Poole, D.; Buchman, D.; Kazemi, S.~M.; Kersting, K.; and Natarajan, S.
\newblock 2014.
\newblock Population size extrapolation in relational probabilistic modelling.
\newblock In {\em Proc. of the Eighth International Conference on Scalable
  Uncertainty Management}.

\bibitem[\protect\citeauthoryear{Poole}{2003}]{Poole:2003}
Poole, D.
\newblock 2003.
\newblock First-order probabilistic inference.
\newblock In {\em Proceedings of the 18th International Joint Conference on
  Artificial Intelligence (IJCAI-03)},  985--991.

\bibitem[\protect\citeauthoryear{Popescul and
  Ungar}{2004}]{popescul2004cluster}
Popescul, A., and Ungar, L.~H.
\newblock 2004.
\newblock Cluster-based concept invention for statistical relational learning.
\newblock In {\em Proceedings of the tenth ACM SIGKDD international conference
  on Knowledge discovery and data mining},  665--670.
\newblock ACM.

\bibitem[\protect\citeauthoryear{Richardson and
  Domingos}{2006}]{Richardson:2006aa}
Richardson, M., and Domingos, P.
\newblock 2006.
\newblock {M}arkov logic networks.
\newblock {\em Machine Learning} 62:107--136.

\bibitem[\protect\citeauthoryear{Schmidt and Murphy}{2010}]{Schmidt:2010}
Schmidt, M.~W., and Murphy, K.~P.
\newblock 2010.
\newblock Convex structure learning in log-linear models: Beyond pairwise
  potentials.
\newblock In {\em International Conference on Artificial Intelligence and
  Statistics}.

\bibitem[\protect\citeauthoryear{Schmidt, Fung, and
  Rosales}{2007}]{schmidt2007fast}
Schmidt, M.; Fung, G.; and Rosales, R.
\newblock 2007.
\newblock Fast optimization methods for l1 regularization: A comparative study
  and two new approaches.
\newblock In {\em Machine Learning: ECML 2007}. Springer.
\newblock  286--297.

\bibitem[\protect\citeauthoryear{Schulte and
  Khosravi}{2012}]{schulte2012learning}
Schulte, O., and Khosravi, H.
\newblock 2012.
\newblock Learning graphical models for relational data via lattice search.
\newblock {\em Machine Learning} 88(3):331--368.

\bibitem[\protect\citeauthoryear{Srinivasan}{2001}]{srinivasan2001aleph}
Srinivasan, A.
\newblock 2001.
\newblock The aleph manual.

\end{thebibliography}
\bibliographystyle{aaai}

\end{document}